\documentclass[journal,11pt,draftclsnofoot,onecolumn]{IEEEtran}

\usepackage{latexsym}
\usepackage{amsfonts}
\usepackage{graphicx} 
\usepackage{subfigure}

\usepackage{amssymb}
\usepackage{natbib}
\usepackage{enumerate}
\usepackage{amsmath}

\usepackage{tabularx}
\usepackage{bbm}

\usepackage{algorithm}
\usepackage{algorithmic}

\usepackage[usenames]{color}

\newcommand{\argmin}[1]{\underset{#1}{\arg\min}}
\newcommand{\argmax}[1]{\underset{#1}{\arg\max}}

\def\BibTeX{{\rm B\kern-.05em{\sc i\kern-.025em b}\kern-.08em
    T\kern-.1667em\lower.7ex\hbox{E}\kern-.125emX}}

\newcolumntype{Z}{>{\raggedleft\arraybackslash}X}

\begin{document}

%
\title{Blind Analysis of EGM Signals: Sparsity-Aware Formulation\thanks{This work has been partly financed by the Spanish government through the CONSOLIDER-INGENIO 2010 program (COMONSENS project, ref. CSD2008-00010), as well as projects  DEIPRO (TEC2009-14504-C02-01), COSIMA (TEC2010-19545-C04-03), ALCIT (TEC2012-38800-C03-01), COMPREHENSION (TEC2012-38883-C02-01) and DISSECT (TEC2012-38058-C03-01).}}

\author{David Luengo$^{\star}$, Javier V\'ia$^{\dagger}$, Sandra Monz\'on$^{\ddagger}$, Tom Trigano$^{\sharp}$ and Antonio Art\'es-Rodr\'iguez$^{\ddagger}$\vspace*{12pt}\\
{\small $^{\star}$ Department of Circuits and Systems Engineering, Universidad Polit\'ecnica de Madrid, 28031 Madrid (Spain).\\
$^{\dagger}$ Department of Communications Engineering, Universidad de Cantabria, 39005 Santander (Spain).\\
$^{\ddagger}$ Department of Signal Proc. and Communic., Universidad Carlos III de Madrid, 28911 Legan\'es (Spain).\\
$^{\sharp}$ Department of Electrical Engineering, Shamoon College of Engineering, Ashdod (Israel).\vspace*{6pt}\\
E-mails: {\tt david.luengo@upm.es, jvia@gtas.dicom.unican.es, smonzon@tsc.uc3m.es, thomast@sce.ac.il, antonio@tsc.uc3m.es}}}
\maketitle 

\vspace{-1.5cm}

\begin{abstract}

This technical note considers the problems of blind sparse learning and inference of electrogram (EGM) signals under atrial fibrillation (AF) conditions.
First of all we introduce a mathematical model for the observed signals that takes into account the multiple foci typically appearing inside the heart during AF.
Then we propose a reconstruction model based on a fixed dictionary and discuss several alternatives for choosing the dictionary.
In order to obtain a sparse solution that takes into account the biological restrictions of the problem, a first alternative is using LASSO regularization followed by a post-processing stage that removes low amplitude coefficients violating the refractory period characteristic of cardiac cells.
As an alternative we propose a novel regularization term, called cross products LASSO (CP-LASSO), that is able to incorporate the biological constraints directly into the optimization problem.
Unfortunately, the resulting problem is non-convex, but we show how it can be solved efficiently in an approximated way making use of successive convex approximations (SCA).
Finally, spectral analysis is performed on the clean activation sequence obtained from the sparse learning stage in order to estimate the number of latent foci and their frequencies.
Simulations on synthetic and real data are provided to validate the proposed approach.

\end{abstract}
\newpage
{\small}

\section{Introduction}
\label{sec:introduction}

The clinical term \emph{atrial fibrillation} (AF) refers to a family of common heart disorders characterized by fast and uncoordinated activations in the atrium.
The mechanisms causing the initiation and maintenance of AF comprise a set of heterogenous interactions at different levels (cells, tissues and the whole heart) changing along time and resulting into different states of AF \citep{Nattel_AFbasicMechanisms2000,Everett_AFbasicMechanisms2004,Nattel_AFmechanismsAnimalModels2005}.
Several theories about the physiological causes underlying AF initiation and maintenance have been formulated over the last 50 years \citep{Nattel_AFtheories2002}.
One of the most prominent hypothesis considers multiple uncoordinated activation foci placed at different locations inside the atrium.
These fast and asynchronous activations cause a disordered global electrical activity that contributes to AF maintenance.   \citep{Krummen_AFinitMechanisms2009}.
In contrast, during normal heart operation conditions (\emph{sinus rhythm}) we observe a single activation focus, placed at the sinus node, acting as a pacemaker for the whole heart and leading to a regular global electrical activity.


In order to understand the pathophysiology of AF, dominant frequency analysis (DFA) has been traditionally used to analyze the data collected from electrocardiograms (ECGs) or electrograms (EGMs).
DFA is useful for identifying the areas corresponding to the highest activation frequencies that may be the drivers maintaining AF, and therefore the targets of ablation therapy for AF termination \citep{Sanders_spectralAnalysis2005}.
However, DFA provides very limited information about the signal's structure, since it is based on the implicit assumption that the underlying signal consists of a single quasi-periodic component plus an irregular component \citep{Barquero_FurierOrganizationAnalysis2010}.
Hence, the only spectral parameter required is the dominant frequency (DF), which tries to characterize the periodicity of the signal, but is very sensitive to distortions and often provides misleading information \citep{Ng_DFtechnicalIssues2007}.

More recently, organization analysis techniques have been introduced, and additional parameters, such as the regularity index (RI) and the organization index (OI), have been used to describe the signals \citep{Fischer_dominantFrequency2007,Barquero_FurierOrganizationAnalysis2010}.
Many other linear and non-linear measures have been proposed for the characterization of AF \citep{Mainardi_NonLinearCoupling2001,Nguyen_AutomatedLocation2010}: the cross-correlation index, the non-linear association measure, the fractionation index, etc.
However, all of them are still based on the same implicit assumption: the observed signals can be modelled by a single regular component plus distortion and noise.


In this technical report we summarize the formulation introduced in \citep{monzon_mlsp_2012} and introduce a novel formulation based on a new sparse regularization term that incorporates the biological restrictions imposed by the refractory period of cardiac cells \citep{luengo_icassp_2013}.
Overall, in these two papers we make two main contributions.
First of all, we introduce a more realistic mathematical model that takes into account the multiple activation foci, and use it to perform spectral analysis, detecting the number of foci and their frequencies.
And secondly, recognizing the sparse nature of the recorded signals, we apply a sparsity-aware learning technique, based on LASSO, to obtain an activation sequence on which the spectral analysis is performed.
In \citep{monzon_mlsp_2012} this is followed by an additional stage that gets rid of spikes that violate the biological restrictions, whereas in \citep{luengo_icassp_2013} we include this term inside the regularization, obtaining a novel regularization term, called cross-products LASSO (CP-LASSO), since it is based on cross-products of coefficients associated to different time instants in the reconstruction model, that we add to the $L_1$ norm regularization term introduced by LASSO.

In the sequel we use bipolar intracardiac electrograms (EGMs), obtained placing a set of electrodes in direct contact with the heart muscle during heart surgery \citep{Ng_DFreview2007,Sanders_spectralAnalysis2005}.
The resulting signal processing algorithm applied to the signals consists of four steps:
\begin{enumerate}
	\item Pre-processing to eliminate potential artifacts, especially outside of the frequency range of interest.
	\item Inferring the spike trains associated to the activation times using a sparsity-aware learning technique based on LASSO \citep{tibshirani_regression_1996}, plus
		a later stage to ensure that biological restrictions are met \citep{monzon_mlsp_2012}, or on CP-LASSO \citep{luengo_icassp_2013} without any additional stage.
	\item Sparse spectral analysis of that activation sequence, using an iterative deflation approach to detect the number of foci and their frequencies.
	\item Post-processing in order to eliminate harmonics and subharmonics.
\end{enumerate}


The report is structured as follows.
First of all, in Section \ref{sec:background} we briefly review the prevalent approach for the analysis of EGMs: dominant frequency analysis.
Then, in Section \ref{sec:problem} we describe the problem formulation used throughout the paper, showing the novel mathematical model (based on a set of unobserved latent signals) proposed for describing the recorded EGMs, the sparsity-aware formulation introduced for solving it, and several dictionaries considered for modelling the unknown latent signals.
Section \ref{sec:mlsp2012} describes the approach proposed in \citep{monzon_mlsp_2012} for inferring the sparse activations: a sparsity-aware learning technique based on LASSO, plus a second stage to incorporate the biological constraints.
The alternative formulation proposed in \citep{luengo_icassp_2013}, based on adding a new regularization term (CP-LASSO) to the sparse learning problem that takes into account the biological constraints, is described in Section \ref{sec:icassp2013}.
Unfortunately, this new regularization term leads to a non-convex optimization problem, so we have to look for methods that are available to produce approximate solutions in a reasonable computational time.
The method chosen, successive convex approximations (SCA), is also described in this section.
Then, Section \ref{sec:SSA} shows how the sparse spike train inferred using either of these two approaches can be used to perform sparse spectral analysis (SSA), thus inferring the number of latent foci as well as their activation frequencies.
%
%
Finally, the conclusions and future lines close the paper in Section \ref{sec:conclusions}.

\section{Background}
\label{sec:background}

\subsection{Dominant Frequency Analysis}
\label{sec:DFA}

Dominant frequency analysis (DFA) is the prevalent approach for the analysis of EGMs.
DFA assumes implicitly that the observed signals are composed of a single regular component (i.e., a quasi-periodic signal) plus an irregular component including the remaining noise and distortion.
Hence, from a mathematical point of view, the $q$-th output (EGM), $1 \le q \le Q$ with $Q$ denoting the number of outputs, can be modelled as \citep{Fischer_dominantFrequency2007}
\begin{equation}
	y_q(t) = \sum_{k=-\infty}^{\infty}{\phi_q(t-k\widetilde{T}_q-\tilde{\tau}_q)} + w_q(t),
\label{eq:DFAoutputModel}
\end{equation}
where $\phi_q(t)$ indicates the average shape of the regular component of the signal, with $\widetilde{T}_q$ denoting its period and $\tilde{\tau}_q$ the delay for $k=0$, and $w_q(t)$ is used to represent the irregular components.
The goal of DFA is characterizing that quasi-periodic signal through its average period, $\widetilde{T}_q$, or equivalently its average frequency, $\tilde{f}_q = 1/\tilde{T}_q$, which is the so called \emph{dominant frequency} (DF).
Occasionally other parameters, such as the organization or the regularity indexes, are also obtained to determine whether the estimated DF is reliable or not \citep{Barquero_FurierOrganizationAnalysis2010,Fischer_dominantFrequency2007}.

The DF is usually obtained separately for each channel using standard spectral analysis techniques.
The typical signal processing approach includes the five steps for each EGM \citep{Fischer_dominantFrequency2007} shown in Algorithm \ref{alg:dfa}.
Several segments can be averaged in order to improve the estimation of the dominant frequency.
However, the ability of the DF to reflect the average atrial activation rate depends on the accuracy of \eqref{eq:DFAoutputModel} in representing the true observed signal.
Unfortunately, several characteristics of atrial activation, such as the complexity of the electrogram morphology, can alter the power spectrum.
In these cases, the DF, $\tilde{f}_q$, is often more related to the complexity of the signal than to the atrial activation rate, thus providing misleading information \citep{Ng_DFtechnicalIssues2007}.

\begin{algorithm}[!htb]
	\begin{enumerate}
		\item[1.] Band-Pass filtering from 30 Hz to 400 Hz.
		\item[2.] Rectification of the resulting signal, recovering near direct current (DC) spectral components.
		\item[3.] Low-Pass filtering with a cut-off frequency of 15 Hz.
		\item[4.] Computation of the spectrum using a localized Fast Fourier Transform (FFT) with a Hanning window of $\Lambda = 4$ s 
			duration, resulting in a resolution $f_{\Lambda} = 1/\Lambda = 0.25$ Hz in the frequency domain.
		\item[5.] Search for the peak with the maximum amplitude in the frequency domain. The frequency associated to this peak is the 
			dominant frequency (DF) of the $q$-th EGM, $\tilde{f}_q$.
	\end{enumerate}
\caption{Dominant frequency analysis (DFA) for the $q$-th signal.}
\label{alg:dfa}
\end{algorithm}

\section{Problem Formulation}
\label{sec:problem}

In this section we show the novel problem formulation proposed as an alternative to the DFA formulation shown in the previous section.
First of all, we introduce a more realistic mathematical model based on the assumption that the observed signals are the result of several unobserved latent functions (the unknown activation foci that we want to estimate) propagating through the heart.
Then, since the real shapes of these latent signals are not precisely known, we introduce a sparsity-aware formulation to solve the problem based on an overcomplete dictionary.

\subsection{Signal Model}
\label{sec:model}

In this technical report we focus on the analysis of electrograms, although the proposed approach can also be applied to other types of signals, as shown in \citep{luengo_icassp_2013}.
Our basic assumption is that the recorded EGMs are composed of the sum of several periodic or quasi-periodic signals plus distortion and noise.
Each of these observed periodic signals are the result of a set of sparse activation foci (spike trains) that propagate through the atrium and reach the sensors.
Hence, these unobserved activations play the role of latent signals, providing a principled way of describing the correlation between the outputs.
Our primary goal here is detecting the number of activation foci, as well as their frequencies.

From a mathematical point of view, let us consider a model with $Q$ correlated outputs, $y_q(t)$, obtained from a set of bipolar electrodes.
These observations are generated by $R$ activation foci (latent signals) propagating inside the atrium, plus noise and interference.
Hence, we model the output of the $q$-th channel ($1 \le q \le Q$) as
\begin{equation}
	y_q(t) = \sum_{r=1}^{R}{p_r(t) * h_{r,q}(t)} + w_q(t),
\label{eq:outputModel}
\end{equation}
where $p_r(t)$ ($1 \le r \le R$) denotes the $r$-th foci, $w_q(t)$ models all the elements in the $q$-th output that cannot be explained by the model (i.e., noise, interferences and distortion), $h_{r,q}(t)$ is the impulse response of the channel between the $r$-th foci and the $q$-th output EGM and $*$ denotes the standard linear convolution operator.\footnote{Note that $h_{r,q}(t)$ includes the response of the sensor and can be slowly time-varying. However, since the sparse learning and the subsequent spectral analysis are performed using short time windows, we can consider the channel to be time-invariant in practice.}
Since we are not interested in recovering the precise shape of the activations, but only in their number and frequencies, we model them as periodic spike trains,\footnote{Note that this is not a limitation, since we can always include the shape of the activations inside the channel's impulse response, $h_{r,q}(t)$. We also remark that the amplitude term, $A_{r,q}[k]$, was not present in the MLSP formulation \citep{monzon_mlsp_2012}. However, we include it here since it allows us to take into account effects such as the amplitude modulation often observed in EGMs or the fact that some activations may not actually be observed (due to blocking phenomena inside the heart, the refractory period of cardiac cells or some other factor).}
\begin{equation}
	p_r(t) = \sum_{k=-\infty}^{\infty}{A_{r,q}[k] \delta(t-kT_r-\tau_r)},
\label{eq:spikeTrains}
\end{equation}
with $\delta(t)$ denoting Dirac's delta, $T_r = 1/f_r$ the average period of the $r$-th spike train (with $f_r$ denoting its associated average frequency) and $\tau_r$ its shift w.r.t. the origin ($0 \le \tau_r < T_r$).\footnote{Let us remark that only a reduced frequency range is meaningful from a physiological point of view. On the one hand, for \emph{sinus rhythm} the heart rate can vary between 30 beats per minute (bpm) and 120 bpm with a typical range of 50--100 bpm, i.e. the range of valid frequencies is $0.5 \le f_r \le 2$ Hz or equivalently $0.5 \le T_r \le 2$ s, with typical ranges $5/6 \le f_r \le 5/3$ Hz or equivalently $0.6 \le T_r \le 1.2$ s. On the other hand, when we analyze EGMs measured during atrial fibrillation (AF), atrial cells can fire at rates of 120--600 bpm (with a typical range of 400--600 bpm) \citep{Nattel_AFtheories2002}, leading to a useful frequency range $2 \le f_r \le 10$ Hz or equivalently $0.1 \le T_r \le 0.5$ s with typical ranges $20/3 \le f_r \le 10$ Hz or equivalently $0.1 \le T_r \le 0.15$ s. Hence, those will be the ranges considered in the sequel: $0.5 \le f_r \le 2$ Hz for sinus rythm and $2 \le f_r \le 10$ Hz for AF.}
Finally, substituting \eqref{eq:spikeTrains} into \eqref{eq:outputModel}, the $q$-th output becomes
\begin{equation}
	y_q(t) = \sum_{r=1}^{R}{\sum_{k=-\infty}^{\infty}{A_{r,q}[k] h_{r,q}(t-kT_r-\tau_r)}} + w_q(t).
\label{eq:outputModel2}
\end{equation}
The discrete-time version of this model, obtained assuming a uniform sampling frequency, $f_s = 1/T_s = 977$ Hz,\footnote{Since the sampling frequency ($f_s = 977$ Hz) is very large compared to the frequencies of interest ($f_r \le 10$ Hz), for simulation purposes we often apply a decimation to the EGM signals, thus obtaining a final sampling frequency $\tilde{f}_s = f_s/L$ with $L \in \{1,2,3,4\}$ (i.e., $244.25 \le \tilde{f}_s(\textrm{Hz}) \le 977$). This allows us to reduce the computational cost of the signal processing algorithms applied without compromising their performance.} would be
\begin{equation}
	y_q[n] = y_q(n T_s) = \sum_{r=1}^{R}{\sum_{k=-\infty}^{\infty}{A_{r,q}[k] h_{r,q}[n,k]}} + w_q[n],
\label{eq:outputModelDiscreteTime}
\end{equation}
where $h_{r,q}[n,k] = h_{r,q}(nT_s-kT_r-\tau_r)$ is the discrete-time equivalent channel and $w_q[n] = w_q(nT_s)$ are the noise plus distortion and interference samples at the sampling instants.\footnote{Note that, due to the discretization, the discrete-time equivalent channel may be time-varying even when $h_{r,q}(t)$ is time-invariant. This is due to a fractional sampling effect, caused by the fact that
\begin{equation}
	h_{r,q}[n,k] = h_{r,q}(nT_s-kT_r-\tau_r) = h_{r,q}\left(\left(n-\frac{kT_r}{T_s}-\frac{\tau_r}{T_s}\right)T_s\right),
\label{eq:discreteTimeChannel}
\end{equation}
and the samples associated to different time-shifts of the channel will not coincide whenever $T_r/T_s$ is not an integer number. Hence, the discrete-time equivalent channel, $h_{r,q}[n,k]$, can indeed be time-varying even when the underlying continuous-time channel, $h_{r,q}(nT_s-kT_r-\tau_r)$, is time-invariant. However, if we assume that $T_s \ll \max_r T_r$ (i.e., $f_s \gg \max_r f_r$), as it occurs in this case, where we have $f_s \ge 244.25$ Hz and $f_r \le 10$ Hz, this effect will be small and we may ignore it.}
In the sequel we make use of this discrete-time model, focusing on inferring the global spike train (i.e., the spike train resulting from the sum of the $R$ foci), and using it to estimate $R$ and $f_r=1/T_r$ for $r=1,\ \ldots,\ R$.

\subsection{Reconstruction Model Based on an Overcomplete Dictionary}
\label{sec:formulationSparseMlsp2012}

Let us denote the $(N+1) \times 1$ vector with the samples from the $q$-th EGM by $\mathbf{y}_q = [y_q[0],\ y_q[1],\ \ldots,\ y_q[N]]^{\top}$, with $y_q[n] = y_q(nT_s)$ obtained sampling $y_q(t)$ uniformly with a sampling frequency $f_s = 1/T_s$ Hz.
Now, let us define the $N \times 1$ vector containing the discrete-time differentiation of the $q$-th output, $\mathbf{z}_q = [z_q[1],\ z_q[2],\ \ldots,\ z_q[N]]^{\top}$ with $z_q[n] = y_q[n] - y_q[n-1]$ for $1 \le n \le N$.
Since we are not interested in the precise shape of the activations, and the number of latent foci is still unknown, we approximate $z_q[n]$ by a mixture of shifted smooth generic curves:\footnote{A detailed analysis of the limits for the convolution in \eqref{eq:diff_model} can be seen in the Appendix.}
\begin{align}
	z_q[n] & = \sum_{m=1}^{M}{\beta_{m,q}[n] * G_m[n]} + \sigma_q \varepsilon_q[n] \nonumber \\
		& = \sum_{m=1}^{M}{\sum_{k=1}^{N}{\beta_{m,q}[k] G_m[n-k]}} + \sigma_q \varepsilon_q[n],
\label{eq:diff_model}
\end{align}
where $\varepsilon_q[n]$ is additive white Gaussian noise (AWGN) with zero-mean and unit variance (i.e., $\varepsilon_q[n] \sim \mathcal{N}(0,1)$), $\sigma_q$ denotes the actual noise variance, assumed to be known or estimated from the data, and $\beta_{m,q}[k]$ is the coefficient of the $q$-th output associated to the $k$-th shift of the $m$-th activation shape, $G_m(t)$, for $1 \le m \le M$, $1 \le q \le Q$ and $1 \le k \le N$.
Note that this model is similar to the discrete-time equivalent model assumed for the data, given by \eqref{eq:outputModelDiscreteTime}, and results in the following equivalent continuous-time model: 
\begin{equation}
	z_q(t) = \sum_{m=1}^{M}{\beta_{m,q}(t) * G_m(t)} + \sigma_q \varepsilon_q(t).
\label{eq:diff_model_time}
\end{equation}
Indeed, the models assumed by the sparsity-aware formulation, given by \eqref{eq:diff_model} and \eqref{eq:diff_model_time}, are very similar to the assumed underlying models, given by \eqref{eq:outputModel} and \eqref{eq:outputModelDiscreteTime}, although there are two important differences:
\begin{enumerate}
	\item Focusing on the discrete-time models, we notice that they describe the first-order time-difference of the sampled EGM signals 
		instead of the signals themselves. Regarding the equivalent continuous-time models, this is akin to working with the first 
		derivative of the signals (which is related to the time-difference in the limit) instead of the signals. This is a common approach 
		to remove the baseline of the signals.
	\item Since the number of activations and their shapes (i.e., the impulse responses associated to the $Q$ channels) are unknown, we 
		use a set of $M \gg R$ activations constructed using generic smooth curves, $G_m(t)$. Note that the same activation shapes are used 
		for all the channels, and we let them select which activations are actually relevant in each case through a sparse learning process 
		based on LASSO or CP-LASSO. This allows us to effectively remove the subindex $q$ from the original activations, $h_{r,q}(t)$,
		moving it to the set of coefficients, $\beta_{m,q}(t)$.
\end{enumerate}

\section{Overcomplete Dictionaries for Sparse Learning}
\label{sec:dictionaries}

In this section we describe several possible choices for the elements of the overcomplete dictionary used in the reconstruction model.

\subsection{Gaussian Dictionary}
\label{sec:gaussianDictionary}

In \citep{monzon_mlsp_2012}, the activation shapes were modelled as samples from truncated and time-shifted Gaussian functions,\footnote{Note that we consider an energy-normalized Gaussian instead of the standard unnormalized Gaussian used in \citep{monzon_mlsp_2012}. The derivation of the normalized Gaussian can be seen in the Appendix.}
\begin{equation}
	\phi_m^{(0)}(t) = G_{m}(t) = \frac{1}{\pi^{1/4}\sqrt{\sigma_m}} \exp\left(-\frac{t^2}{2\sigma_m^2}\right) \qquad \textrm{for} \quad -T_m \le t \le T_m,
\label{eq:gauss_base}
\end{equation}
with $T_m = N_m T_s$ a user-defined threshold (set up in practice so that $G_m(\pm T_m)$ is close to zero), and $\sigma^2_m$ a finite set of $M \ge R$ user-defined variances with $\sigma_1^2 < \sigma_2^2 < \ldots <
\sigma_M^2$.
Let us define $\displaystyle{T_{\max} = \max_{1 \le m \le M}\ T_m = T_M}$ and $\displaystyle{N_{\max} = \max_{1 \le m \le M}\ N_m = N_M}$, with $N_m = \lfloor T_m/T_s \rfloor$ and $\lfloor x \rfloor$ denoting the integer part of the real number $x$.
Now, from the continuous-time activation shape, $G_m(t)$ with support $-T_M \le t \le T_M$,\footnote{We have to consider the largest support for all the activation shapes in the dictionary, even though we know that $G_m(t)=0$ for $|t| > T_m$.} we can construct the discrete-time activation elements through uniform sampling with a period $T_s=1/f_s$ and time-shifting by $N_M$ samples, i.e.,
\begin{equation}
	\phi_m^{(0)}[n] = G_m[n] = G_m((n-N_M)T_s) = G_m((n-\lfloor T_M/T_s \rfloor)T_s).
\label{eq:discreteTimeActivations}
\end{equation}
Hence, all the discrete-time activation elements suffer a delay of $N_M$ samples (i.e., $N_M T_s$ seconds) that must be taken into account when interpreting the results obtained.

Now we can rewrite the sparse model in \eqref{eq:diff_model} more compactly in matrix form by defining a set of $N \times M$ matrices, $\mathbf{\Phi}_k$ for $1 \le k \le N$, such that their $(n,m)$-th element is $\mathbf{\Phi}_k(n,m) = \phi_m^{(0)}[n-k] = G_{m}[n-k]$ for $1 \leq n \leq N$ and $1 \leq m \leq M$, i.e.,
\begin{equation}
	\mathbf{\Phi}_k = \left[
		\begin{array}{cccc}
			G_1[1-k] & G_2[1-k] & \cdots & G_M[1-k] \\
			G_1[2-k] & G_2[2-k] & \cdots & G_M[2-k] \\
			\vdots & \vdots & & \vdots \\
			G_1[n-k] & G_2[n-k] & \cdots & G_M[n-k] \\
			\vdots & \vdots & & \vdots \\
			G_1[N-k] & G_2[N-k] & \cdots & G_M[N-k]
		\end{array}
		\right].
\end{equation}
Concatenating all these matrices we obtain an overcomplete global dictionary (note that we have $MN$ dictionary elements and only $N < MN$ samples) that can be collected in the following $N \times MN$ matrix,\footnote{Note that, due to the use of truncated and time-shifted Gaussians, $G_m[n]=0$ whenever $n < 0$ or $n > 2N_m$. Hence, many elements in $\mathbf{\Phi}_k$ ($1 \le k \le N$), and thus also in $\mathbf{\Phi}$, will actually be zero, as sketched in \eqref{eq:dictionary}.}
\begin{align}
	\mathbf{\Phi} & = [\mathbf{\Phi}_1,\ \mathbf{\Phi}_2,\ \ldots, \ \mathbf{\Phi}_N] \nonumber \\
		& = \left[
		\begin{array}{cccccccccc}
			G_1[0] & \cdots & G_M[0] & G_1[-1] & \cdots & G_M[-1] & \cdots & G_1[1-N] & \cdots & G_M[1-N] \\
			G_1[1] & \cdots & G_M[1] & G_1[0] & \cdots & G_M[0] & \cdots & G_1[2-N] & \cdots & G_M[2-N] \\
			\vdots &  & \vdots & \vdots &  & \vdots & & \vdots &  & \vdots \\
			G_1[n-1] & \cdots & G_M[n-1] & G_1[n-2] & \cdots & G_M[n-2] & \cdots & G_1[n-N] & \cdots & G_M[n-N] \\
			\vdots &  & \vdots & \vdots &  & \vdots & & \vdots &  & \vdots \\
			G_1[N-1] & \cdots & G_M[N-1] & G_1[N-2] & \cdots & G_M[N-2] & \cdots & G_1[0] & \cdots & G_M[0] \\
		\end{array}
		\right] \nonumber \\
		& = \left[
		\begin{array}{cccccccccc}
			G_1[0] & \cdots & G_M[0] & 0 & \cdots & 0 & \cdots & 0 & \cdots & 0 \\
			G_1[1] & \cdots & G_M[1] & G_1[0] & \cdots & G_M[0] & \cdots & 0 & \cdots & 0 \\
			\vdots &  & \vdots & \vdots &  & \vdots & & \vdots &  & \vdots \\
			G_1[n-1] & \cdots & G_M[n-1] & G_1[n-2] & \cdots & G_M[n-2] & \cdots & 0 & \cdots & 0 \\
			\vdots &  & \vdots & \vdots &  & \vdots & & \vdots &  & \vdots \\
			0 & \cdots & 0 & 0 & \cdots & 0 & \cdots & G_1[0] & \cdots & G_M[0] \\
		\end{array}
		\right],
\label{eq:dictionary}
\end{align}
where we have assumed that $N \gg 2N_M$ in the last expression, as is usually the case in practice.
Now, using \eqref{eq:dictionary}, \eqref{eq:diff_model} can be expressed in a completely equivalent way as
\begin{equation}
	\mathbf{z}_q = \mathbf{\Phi} \boldsymbol\beta_q + \sigma_q \boldsymbol\varepsilon_q,
\label{eq:regression}
\end{equation}
where $\boldsymbol\varepsilon_q = [\varepsilon_q[1],\ \varepsilon_q[2],\ \ldots,\ \varepsilon_q[N]]^{\top}$ is an $N \times 1$ column vector with the noise samples associated to each sample of $z_q[n]$, and  $\boldsymbol\beta_q$ is an $MN \times 1$ column vector composed of $N$ subvectors of size $M$:
\begin{align}
	\boldsymbol\beta_q & = [\boldsymbol\beta_{q}^{\top}[1],\ \boldsymbol\beta_{q}^{\top}[2],\ \ldots,\ 
		\boldsymbol\beta_{q}^{\top}[N]]^{\top},\label{eq:fullCoefficientVector}\\
	\boldsymbol\beta_{q}[k] & = [\beta_{1,q}[k],\ \ldots,\ \beta_{M,q}[k]]^{\top},\ 1 \le k \le N.
	\label{eq:partialCoefficientVector}
\end{align}

Finally, note that this dictionary is not fitted to detect activation times close to the initial boundary of the signal (i.e., $n=1$).
However, this issue can be easily circumvented by adding $2N_M$ zeros to the signal to be processed before $z[1]$.\footnote{An alternative way of avoiding this problem is by guaranteeing that no activation is present in the observations inside the first $N_M$ samples.}
This would result in an extended support for the sequence, $-(2N_M-1) \le n \le N$, an extended discrete-time differentiation vector,
\begin{equation}
	\widetilde{\mathbf{z}}_q[n] = [\underbrace{0,\ \ldots,\ 0}_{2N_M\ \textrm{zeros}},\
		\underbrace{z_q[1],\ \ldots,\ z_q[N]}_{N\ \textrm{samples}}]^{\top},
\label{eq:extendedSampleVector}
\end{equation}
an extended coefficients vector,
\begin{align}
	\widetilde{\boldsymbol\beta}_q & = [\boldsymbol\beta_{q}^{\top}[-(2N_M-1)],\ \ldots,\ \boldsymbol\beta_{q}^{\top}[-1],\ 
		\boldsymbol\beta_{q}^{\top}[0],\ \boldsymbol\beta_{q}^{\top}[1],\ \ldots,\ 
		\boldsymbol\beta_{q}^{\top}[N]]^{\top},
	\label{eq:extendedCoefficientVector}
\end{align}
with $\boldsymbol\beta_{q}^{\top}[k]$ still given by \eqref{eq:partialCoefficientVector} for $-(2N_M-1) \le k \le N$, and an extended dictionary matrix,
\begin{equation}
	\widetilde{\mathbf{\Phi}} = [\widetilde{\mathbf{\Phi}}_{-(2N_M-1)},\ \ldots,\ \widetilde{\mathbf{\Phi}}_{-1},\
		\widetilde{\mathbf{\Phi}}_0,\ \widetilde{\mathbf{\Phi}}_1,\ \widetilde{\mathbf{\Phi}}_2,\ \ldots, \ \widetilde{\mathbf{\Phi}}_N],
\label{eq:extendedDictionary}
\end{equation}
with 
\begin{equation}
	\widetilde{\mathbf{\Phi}}_k = \left[
		\begin{array}{cccc}
			G_1[-(2N_M-1)-k] & G_2[-(2N_M-1)-k] & \cdots & G_M[-(2N_M-1)-k] \\
			\vdots & \vdots & & \vdots \\
			G_1[0-k] & G_2[0-k] & \cdots & G_M[0-k] \\
			G_1[1-k] & G_2[1-k] & \cdots & G_M[1-k] \\
			\vdots & \vdots & & \vdots \\
			G_1[N-k] & G_2[N-k] & \cdots & G_M[N-k]
		\end{array}
		\right]
\label{eq:extendedDictionarySubmatrix}
\end{equation}
again for $-(2N_M-1) \le k \le N$.

\subsection{Ideal Dictionary}
\label{sec:idealDictionary}

The optimum dictionary would in fact be composed of a set of $Q$ dictionaries tailored to the characteristics of each of the $Q$ outputs.
More specifically, the dictionary for the $q$-th output would be given by the following $N \times RN$ matrix,
\begin{equation}
	\mathbf{\Phi}_q = [\mathbf{\Phi}_{1,q},\ \mathbf{\Phi}_{2,q},\ \ldots, \ \mathbf{\Phi}_{N,q}],
\label{eq:idealDictionary}
\end{equation}
where $\mathbf{\Phi}_{k,q}$ is the $N \times R$ matrix composed of the $R$ impulse responses between the $r$-th latent signal ($1 \le r \le R$) and the $q$-th observation, i.e., 
\begin{equation}
	\mathbf{\Phi}_{k,q} = \left[
		\begin{array}{cccc}
			h_{1,q}[1-k] & h_{2,q}[1-k] & \cdots & h_{R,q}[1-k] \\
			h_{1,q}[2-k] & h_{2,q}[2-k] & \cdots & h_{R,q}[2-k] \\
			\vdots & \vdots & & \vdots \\
			h_{1,q}[n-k] & h_{2,q}[n-k] & \cdots & h_{R,q}[n-k] \\
			\vdots & \vdots & & \vdots \\
			h_{1,q}[N-k] & h_{2,q}[N-k] & \cdots & h_{R,q}[N-k]
		\end{array}
		\right].
\end{equation}
Using this dictionary, \eqref{eq:diff_model} can be expressed now as
\begin{equation}
	\mathbf{z}_q = \mathbf{\Phi}_q \boldsymbol\beta_q + \sigma_q \boldsymbol\varepsilon_q,
\label{eq:idealRegression}
\end{equation}
where $\boldsymbol\beta_q$ still has the structure described in \eqref{eq:fullCoefficientVector}, with $\boldsymbol\beta_{q}[k] = [\beta_{1,q}[k],\ \ldots,\ \beta_{R,q}[k]]^{\top}$.
This dictionary would lead to the sparsest possible solution, consisting in approximately $NQ\sum_{r=1}^{R}{T_s/T_r}$ non-zero elements that will coincide with the amplitudes of the activations, $A_{r,q}[k]$.

\subsection{Alternative Dictionaries}
\label{sec:alternativeDictionaries}

Unfortunately, the ideal dictionary discussed in the previous section requires either knowledge of the $h_{r,q}(t)$ or a reliable estimation, something which is not easy for the application considered.
However, it provides us with a criterion for comparing different dictionaries: the best dictionary will be the one that attains the sparsest representation (thus allowing us to get closer to the lower bound provided by the unavailable ideal dictionary), while obtaining a good reconstruction error (e.g., ensuring that the $L_2$ norm of the reconstruction error is below a given threshold).
We are currently considering better dictionaries using wavelets or wavelet packets, and even activations extracted from real data, since this should lead to sparser solutions with a good reconstruction error.
As an example, in \citep{luengo_icassp_2013} we consider the Mexican hat wavelet, also known as Ricker wavelet,\footnote{The derivation of the mexican hat wavelet can be seen in the Appendix.}
\begin{equation}
	\phi_m^{(2)}(t) = \mathcal{R}_m(t) = {2 \over {\pi^{1/4}\sqrt{3\sigma_m}}} \left( 1 - {t^2 \over \sigma_m^2} \right)
		\exp\left(-\frac{t^2}{2\sigma_m^2}\right),
\label{eq:mexicanHat}
\end{equation}
which is the negative normalized second derivative of a Gaussian function, and is used due to its similarity to activations observed in real data.\footnote{We remark that both the Gaussian and the mexican hat belong to the class of \emph{Hermitian wavelets}, so called because the amplitude of the $\ell$-th Hermitian wavelet depends on the $\ell$-th order Hermite polynomial. Indeed, the superscript in $\phi_m^{(\ell)}(t)$ indicates the order of the wavelet (zero for the Gaussian and two for the mexican hat). Further details can be seen in the Appendix.}

\section{Indirect Sparse Solution: LASSO plus Post-Processing}
\label{sec:mlsp2012}

In \citep{monzon_mlsp_2012} we obtained a sparse vector of coefficients for \eqref{eq:regression}, $\boldsymbol\beta_q$, following a two step procedure, which is a variation of the algorithm introduced in~\citep{trigano_nonhomogeneous_2011}: an initial sparse solution obtained  by applying a LASSO regularization is followed by a greedy procedure for selecting only the largest coefficients that respect the biological constraints.

In order to obtain a sparse regressor, from which the information on the arrival times can be retrieved, we estimate $\boldsymbol{\beta}_q$ initially by means of LASSO~\citep{tibshirani_regression_1996}.
Namely, $\hat{\boldsymbol{\beta}}_q^{L_1}(\lambda_q)$ is given by
\begin{equation}
  \hat{\boldsymbol{\beta}}_q^{L_1}(\lambda_q) = \argmin{\boldsymbol\beta_q \in \mathbb{R}^{MN}}
  	\left\{ \frac{1}{2N}  \left \Vert \mathbf{z}_q - \mathbf{A} \boldsymbol\beta_q \right \|_2^2 
  	+ \lambda_q \, \|\boldsymbol\beta_q\|_1 \right\},
\label{eq:Lasso_functional}
\end{equation}
where $\|\boldsymbol\beta_q\|_1$ denotes the $L_1$ norm of $\boldsymbol\beta_q$ and $\lambda_q$ indicates the trade-off between sparsity and estimation precision: the higher the value of $\lambda_q$ the more emphasis will be placed on obtaining a sparse solution, although at the expense of an increased quadratic error in the approximation.\footnote{In \citep{monzon_mlsp_2012}, $\lambda_q = 10^{-4}$ was used for the sinus rhythm simulations and $\lambda_q = 10^{-6}$ for the AF simulations. Note that having a smaller value of $\lambda_q$ for AF implies that a less sparse solution will be obtained for AF in comparison to sinus rhythm. Both values of $\lambda_q$ were obtained through an exhaustive search using real data.}

However, in order to obtain an even sparser representation that takes into account the physiological restrictions imposed on the signals, we introduce an additional step after the computation of $\hat{\boldsymbol{\beta}}_q^{L_1}(\lambda_q)$.
The samples associated to the arrival times of the spikes are estimated recursively as follows:
\begin{align}
	\hat{n}_{k,q} & = \argmax{n=1,\ldots,N} \left\{ \|\hat{\boldsymbol\beta}_q^{L_1}[n]\|_1
		\mathbb{I}(\eta_q < \|\hat{\boldsymbol\beta}_q^{L_1}[n]\|_1
			< \|\hat{\boldsymbol\beta}_q^{L_1}[\hat{n}_{k-1,q}]\|_1)\right\} \nonumber \\
& \textrm{s.t.} \quad |\hat{n}_{k,q}-\hat{n}_{\ell,q}| > N_{\min}, \quad \textrm{for} \quad \ell = 1,\ \ldots,\ k-1,
\label{eq:arrival_estimates}
\end{align}
where $\mathbb{I}(\cdot)$ is an indicator function, i.e., a function that takes a value equal to one if the logical condition is fulfilled and zero otherwise,
\begin{equation}
	\mathbb{I}(\eta_q < \|\hat{\boldsymbol\beta}_q^{L_1}[n]\|_1 < \|\hat{\boldsymbol\beta}_q^{L_1}[\hat{n}_{k-1,q}]\|_1) =
		\begin{cases}
			1, & \eta_q < \|\hat{\boldsymbol\beta}_q^{L_1}[n]\|_1 < \|\hat{\boldsymbol\beta}_q^{L_1}[\hat{n}_{k-1,q}]\|_1;\\
			0, & \textrm{otherwise},
		\end{cases}
\label{eq:indicator}
\end{equation}
and $\eta_q$ and $N_{\min}$ are user-defined thresholds.
The first one, $\eta_q$, is used to discard the $\hat{\boldsymbol\beta}_q^{L_1}[n]$ with a small $L_1$ norm, which contribute to improve the signal reconstruction but provide little information on the localization of the spikes. We have found out empirically that choosing $\eta_q = 3\sigma_q$ provides good results.\footnote{In \citep{monzon_mlsp_2012}, $\sigma_q = 2 \cdot 10^{-3}$ was used for the simulations performed under induced sinus rythm, and $\sigma_q = 5 \cdot 10^{-4}$ for the atrial fibrillation (AF) simulations. Note that, as $\sigma_q^2$ represents the unexplained variance in the reconstruction model (i.e., the energy of the reconstruction error), using a lower value for AF implies allowing less reconstruction error, which in turn results in a less sparse solution. In both cases, these values were obtained through an exhaustive search using real data.}
The second one, $N_{\min}$, accounts for the fact that consecutive pulses cannot overlap. Thus, in practice $N_{\min}$ is chosen in such a way that $N_{\min}/f_s \approx 100\ \text{ms}$ (i.e., $N_{\min} \approx f_s/10$), which is a standard value for the refractory period.\footnote{Since $f_s = 977$ Hz, taking the integer part of $f_s/10$ we obtain $N_{\min} = 97$ when no decimation is applied (i.e., $L=1$), $N_{\min} = 48$ for $L=2$, $N_{\min} = 32$ for $L=3$ and  $N_{\min} = 24$ for $L=4$.}
The final procedure used in practice to implement \eqref{eq:arrival_estimates} for the $q$-th channel is an iterative greedy approach that follows the steps shown in Algorithm \ref{alg:greedySpikeSelection}.

\begin{algorithm}[!htb]
	\begin{enumerate}
		\item[1.] Initialization: set $k=1$ and $\beta_{\max} = \infty$.
		\item[2.] If $\beta_{\max} > \eta_q$:
		\begin{enumerate}
			\item[2.1.] Select the index corresponding to the largest coefficient:
				\begin{equation}
					\hat{n}_{k,q} = \argmax{n=1,\ldots,N}\ \|\hat{\boldsymbol\beta}_q^{L_1}[n]\|_1.
				\label{eq:largestCoef}
				\end{equation}
			\item[2.2.] If $\|\hat{\boldsymbol\beta}_q^{L_1}[\hat{n}_{k,q}]\|_1 \le \eta_q$, then END.
			\item[2.3] Otherwise, check whether $|\hat{n}_{k,q}-\hat{n}_{\ell,q}| > N_{\min}$ for $\ell = 1,\ \ldots,\ k-1$. If this condition
				is fulfilled, store $\hat{n}_{k,q}$ and	$\hat{\boldsymbol\beta}_q^{L_1}[\hat{n}_{k,q}]$,
				set $\beta_{\max} = \|\hat{\boldsymbol\beta}_q^{L_1}[\hat{n}_{k,q}]\|_1$, $k=k+1$,
				$\hat{\boldsymbol\beta}_q^{L_1}[\hat{n}_{k,q}] = 0$
			\item[2.4] Return to step 2.1.
		\end{enumerate}
	\end{enumerate}
\caption{Iterative greedy approach for selection of the final spikes for the $q$-th signal.}
\label{alg:greedySpikeSelection}
\end{algorithm}

Following this procedure we obtain a set of $P$ arrival times and their associated amplitudes,\footnote{Note that this procedure can also be used in practice to discard noisy channels that contain no valid information. Since $P_{\min} \approx \frac{NT_s}{\max_r\ T_r}$, we may automatically discard those channels with $P < P_{\min}$ as invalid, as the activations in those channels will correspond to occasional large noise samples.} that we may use to construct an \emph{activation sequence} (also called spike train) composed of Kronecker deltas at the locations of the activations,\footnote{Note that we have not estimated $R$ yet. Hence, we cannot separate the contribution of each foci to \eqref{eq:cleanSignal} as we did in the original model, given by \eqref{eq:outputModel2}.}
\begin{equation}
	\pi_q[n] = \sum_{k=1}^{P}{\delta[n-\hat{n}_{k,q}]}.
\label{eq:cleanSignal}
\end{equation}
This sequence was used in \citep{monzon_mlsp_2012} to perform a spectral analysis of the clean signal given by \eqref{eq:cleanSignal}, since it allows us to get rid of the effect of the unknown channels, $h_{r,q}(t)$, and the particular dictionary used, given by $G_m(t)$.
Alternatively, we may construct this spike train taking into account the amplitudes associated to each coefficient, i.e.,
\begin{equation}
	\widetilde{\pi}_q[n] = \sum_{k=1}^{P}{\|\hat{\boldsymbol\beta}_q^{L_1}[\hat{n}_{k,q}]\|_1 \delta[n-\hat{n}_{k,q}]}.
\label{eq:cleanSignal2}
\end{equation}
Whether this will provide useful information for the spectral analysis or not remains an open question.

\section{Direct Sparse Solution: Cross-Products LASSO}
\label{sec:icassp2013}

\subsection{One-Step Sparsity-Aware Formulation}
\label{sec:formulationSparseIcassp2013}

Instead of following a standard sparse regression initially using LASSO, as given by \eqref{eq:Lasso_functional}, and then having to perform a further post-processing stage to take into account the biological restrictions of the problem using \eqref{eq:arrival_estimates}, we would like to include the problem's constraints into the sparse formulation.
This would provide us with a more elegant formulation, potentially allowing us to obtain a better solution to the problem and in a more efficient way also.
Note that the cost function used by LASSO is
\begin{equation}
	J_{\textrm{LASSO}} = \frac{1}{2N}  \left \Vert \mathbf{z}_q - \mathbf{A} \boldsymbol\beta_q \right \|_2^2
		+ \lambda_q \, \|\boldsymbol\beta_q\|_1,
\label{eq:costFunctionLasso}
\end{equation}
i.e., it is composed of the least squares (LS) error between the model, $\mathbf{A} \boldsymbol\beta_q$, and the data, $\mathbf{z}_q$, plus a regularization term, $R_{\textrm{LASSO}} = \lambda_q \, \|\boldsymbol\beta_q\|_1$, that enforces a sparsity-aware solution.
On the one hand, the first constraint imposed by the post-processing stage in \citep{monzon_mlsp_2012}, having $|\beta_q[n]| > \eta_q$, can be accommodated by selecting a value of $\lambda_q$ large enough, so it does not require any modification in the cost function.
On the other hand, the second constraint imposed is related to the refractory period associated to cardiac cells, and requires two coefficients, $\beta_{\ell,q}[n]$ and $\beta_{m,q}[n+k]$ for $1 \le \ell, m \le M$, to be zero when the distance between the centers of their associated activation shapes is less than $N_{\min} = \lfloor f_s/10 \rfloor$.
In order to incorporate this restriction to the cost function, we need to add a new regularization term to \eqref{eq:costFunctionLasso} that takes into account this distance restriction.
This can be done using the $L_0$ ``norm'',\footnote{The $L_0$ ``norm'' of a vector, $\|\mathbf{x}\|_0$, is not really a norm, since it does not satify the triangle inequality. Hence, some authors refer to it as a counting function (see e.g. \citep{Tropp2010}). However, with a slight abuse of terminology, here we will refer to it as a ``norm''.} and results in the following modified cost function
\begin{equation}
	J_{\textrm{exact}} = \frac{1}{2N}  \left \Vert \mathbf{z}_q - \mathbf{A} \boldsymbol\beta_q \right \|_2^2
		+ \lambda_q \, \|\boldsymbol\beta_q\|_1
		+ \rho_q \sum_{n=1}^{N}{\sum_{m=1}^{M}{\sum_{\substack{k=-N_{\min} \\ k \ne 0}}^{N_{\min}}
			{\|\beta_{m,q}[n]\boldsymbol\beta_q[n+k]\|_0}}},
\label{eq:costFunctionModExact}
\end{equation}
where $\rho_q$ is an additional regularization parameter, $\|\cdot\|_0$ denotes the $L_0$ ``norm'' of a vector (i.e., the number of non-zero elements), $\beta_{m,q}[n]$ is the coefficient associated to the $m$-th activation shape of the $q$-th channel centered around the $n$-th sample, $\boldsymbol\beta_q[n+k] = [\beta_{1,q}[n+k],\ \ldots,\ \beta_{M,q}[n+k]]^{\top}$ is the $M \times 1$ vector containing the coefficients associated to all the activation shapes of the $q$-th channel centered around the $(n+k)$-th sample, and $N_{\min} = \lfloor f_s/(10L) \rfloor$, with $1 \le L \le 4$ indicating the decimation rate, is the minimum number of zero-valued coefficients required between two consecutive activations due to biological reasons.
Note that the newly introduced regularization term, $\|\beta_{m,q}[n]\boldsymbol\beta_q[n+k]\|_0$, is equal to the number of activations that violate the biological constraint, since the vector resulting from the product,
$\beta_{m,q}[n]\boldsymbol\beta_q[n+k] = [\beta_{m,q}[n]\beta_{1,q}[n+k],\ \ldots,\ \beta_{m,q}[n]\beta_{M,q}[n+k]]^{\top}$, will contain a non-zero term whenever a shape is active simultaneously at the $n$-th and $(n+k)$-th sample for $-N_{\min} \le k \le N_{\min}$ with $k \ne 0$.
Hence, the last regularization term penalizes violations of the biological constraints, and indeed, by letting $\rho_q \to \infty$ (or by taking a very large value in practice, i.e., $\rho_q \gg \lambda_q$ and $\rho_q \gg 1/(2N)$) and choosing an appropriate value of $\lambda_q$, this cost function solves exactly the same problem as the original cost function plus the post-processing stage.

Unfortunately, the $L_0$ ``norm'' is generally intractable, and the general approach taken is substituting it by the more tractable $L_1$ norm, which provides an equivalent solution under certain conditions (often difficult to check in practice).
Performing this standard relaxation, the modified cost function given by \eqref{eq:costFunctionModExact} turns into the following cost function,\footnote{Note that, when moving from \eqref{eq:costFunctionModExact} to \eqref{eq:costFunctionModApprox}, we cannot integrate the two regularization terms into a single one, as the newly introduced term by itself does not lead to a sparse solution.}
\begin{equation}
	J_{\textrm{approx}} = \frac{1}{2N}  \left \Vert \mathbf{z}_q - \mathbf{A} \boldsymbol\beta_q \right \|_2^2
		+ \lambda_q \, \|\boldsymbol\beta_q\|_1
		+ \rho_q \sum_{n=1}^{N}{\sum_{m=1}^{M}{\sum_{\substack{k=-N_{\min} \\ k \ne 0}}^{N_{\min}} 
			{\|\beta_{m,q}[n]\boldsymbol\beta_q[n+k]\|_1}}}.
\label{eq:costFunctionModApprox}
\end{equation}
The additional regularization term in \eqref{eq:costFunctionModApprox} can be expressed alternatively as
\begin{align}
	R_{\textrm{approx}}
	 & = \sum_{n=1}^{N}{\sum_{m=1}^{M}{\sum_{\substack{k=-N_{\min} \\ k \ne 0}}^{N_{\min}} 
	 	{\|\beta_{m,q}[n]\boldsymbol\beta_q[n+k]\|_1}}} \\
	 & = \sum_{n=1}^{N}{\sum_{\substack{k=-N_{\min} \\ k \ne 0}}^{N_{\min}}
	 	{\|\textrm{vec}(\boldsymbol\beta_q[n]\boldsymbol\beta_q^{\top}[n+k])\|_1}} \\
	 & = \sum_{n=1}^{N}{\|\textrm{vec}(\boldsymbol\beta_q[n]
		[\boldsymbol\beta_q^{\top}[n-N_{\min}],\ \ldots,\ \boldsymbol\beta_q^{\top}[n-1],\ \boldsymbol\beta_q^{\top}[n+1],\ 
			\ldots,\ \boldsymbol\beta_q^{\top}[n+N_{\min}]])\|_1} \\
	& = \|\textrm{vec}([\boldsymbol\beta_q[1],\ \boldsymbol\beta_q[2],\ \ldots,\ \boldsymbol\beta_q[N]] \mathbf{B}_q)\|_1
\label{eq:regularizationTerm}
\end{align}
where $\textrm{vec}(\cdot)$ denotes the vectorization of a matrix (i.e., the column vector constructed by stacking all the elements of the matrix taken column by column), $[\boldsymbol\beta_q[1],\ \boldsymbol\beta_q[2],\ \ldots,\ \boldsymbol\beta_q[N]]$ is an $M \times N$ matrix, and the $N \times 2MN_{\min}$ matrix $\mathbf{B}_q$ in the last equation is given by
\begin{align}
	\mathbf{B}_q & = \left[
		\begin{array}{cccccc}
		\boldsymbol\beta_q^{\top}[1-N_{\min}] & \cdots & \boldsymbol\beta_q^{\top}[0] & \boldsymbol\beta_q^{\top}[2] & \cdots &
			\boldsymbol\beta_q^{\top}[1+N_{\min}] \\
		\boldsymbol\beta_q^{\top}[2-N_{\min}] & \cdots & \boldsymbol\beta_q^{\top}[1] & \boldsymbol\beta_q^{\top}[3] & \cdots & 
			\boldsymbol\beta_q^{\top}[2+N_{\min}] \\
		\vdots & & \vdots & & \vdots \\
		\boldsymbol\beta_q^{\top}[N-1-N_{\min}] & \cdots & \boldsymbol\beta_q^{\top}[N-2] & \boldsymbol\beta_q^{\top}[N] & \cdots & 
			\boldsymbol\beta_q^{\top}[N-1+N_{\min}] \\
		\boldsymbol\beta_q^{\top}[N-N_{\min}] & \cdots & \boldsymbol\beta_q^{\top}[N-1] & \boldsymbol\beta_q^{\top}[N+1] & \cdots & 
			\boldsymbol\beta_q^{\top}[N+N_{\min}]
		\end{array}
	\right] \nonumber \\
	& = \left[
		\begin{array}{cccccc}
		\mathbf{0}_M^{\top} & \cdots & \mathbf{0}_M^{\top} & \boldsymbol\beta_q^{\top}[2] & \cdots &
			\boldsymbol\beta_q^{\top}[1+N_{\min}] \\
		\mathbf{0}_M^{\top} & \cdots & \boldsymbol\beta_q^{\top}[1] & \boldsymbol\beta_q^{\top}[3] & \cdots & 
			\boldsymbol\beta_q^{\top}[2+N_{\min}] \\
		\vdots & & \vdots & & \vdots \\
		\boldsymbol\beta_q^{\top}[N-1-N_{\min}] & \cdots & \boldsymbol\beta_q^{\top}[N-2] & \boldsymbol\beta_q^{\top}[N] & \cdots & 
			\mathbf{0}_M^{\top} \\
		\boldsymbol\beta_q^{\top}[N-N_{\min}] & \cdots & \boldsymbol\beta_q^{\top}[N-1] & \mathbf{0}_M^{\top} & \cdots & 
			\mathbf{0}_M^{\top}
		\end{array}
	\right],\label{eq:matrixB}
\end{align}
where the second expression is obtained by noting that $\boldsymbol\beta_q[k] = \mathbf{0}_M$ for $k < 1$ and $k > N$.
Finally, we note that the vectorization of the product of a $k \times \ell$ matrix $\mathbf{A}$ and another $\ell \times m$ matrix $\mathbf{B}$ can be expressed as
\begin{equation}
	\textrm{vec}(\mathbf{AB}) = (\mathbf{I}_m \otimes \mathbf{A}) \textrm{vec}(\mathbf{B})
		= (\mathbf{B}^{\top} \otimes \mathbf{I}_k) \textrm{vec}(\mathbf{A}),
\label{eq:vectorization}
\end{equation}
where $\mathbf{I}_p$ represents the $p \times p$ identity matrix and $\otimes$ denotes the Kronecker product of two matrices \citep{VanLoan2000}.
The Kronecker product of an $m \times n$ matrix $\mathbf{A}$ and a $p \times q$ matrix $\mathbf{B}$, results in the following $mp \times nq$ matrix $\mathbf{C}$:
\begin{equation}
	\mathbf{C} = \left[
		\begin{array}{cccc}
			a_{11} \mathbf{B} & a_{12} \mathbf{B} & \cdots & a_{1n} \mathbf{B} \\
			a_{21} \mathbf{B} & a_{22} \mathbf{B} & \cdots & a_{2n} \mathbf{B} \\
			\vdots & \vdots & \ddots & \vdots \\
			a_{m1} \mathbf{B} & a_{m2} \mathbf{B} & \cdots & a_{mn} \mathbf{B}
		\end{array}
	\right],
\label{eq:kroneckerProduct}
\end{equation}
with $a_{ij}$ denoting the $(i,j)$-th element of matrix $\mathbf{A}$.
Applying \eqref{eq:vectorization} to the last expression of \eqref{eq:regularizationTerm}, we finally get
\begin{align}
	R_{\textrm{approx}} & =
		\|(\mathbf{I}_{2MN_{\min}} \otimes [\boldsymbol\beta_q[1],\ \boldsymbol\beta_q[2],\ \ldots,\ \boldsymbol\beta_q[N]]) 
			\textrm{vec}(\mathbf{B}_q)\|_1 \nonumber \\
		& = \|(\mathbf{B}_q^{\top} \otimes \mathbf{I}_M) \boldsymbol\beta_q\|_1,
\end{align}
where we have used the fact that $\textrm{vec}([\boldsymbol\beta_q[1],\ \boldsymbol\beta_q[2],\ \ldots,\ \boldsymbol\beta_q[N]]) = \boldsymbol\beta_q$ in the last expression, and the cost function that we want to minimize finally becomes
\begin{equation}
	J_{\textrm{approx}} = \frac{1}{2N}  \left \Vert \mathbf{z}_q - \mathbf{A} \boldsymbol\beta_q \right \|_2^2
		+ \lambda_q \| \boldsymbol\beta_q \|_1
		+ \rho_q \|(\mathbf{B}_q^{\top} \otimes \mathbf{I}_M) \boldsymbol\beta_q\|_1.
\label{eq:costFunctionModApprox2}
\end{equation}
Note that the structure of the new regularization term added, $R_{\textrm{approx}} = \rho_q \|(\mathbf{B}_q^{\top} \otimes \mathbf{I}_M) \boldsymbol\beta_q\|_1$, is conditioned by the Kronecker product between $\mathbf{B}_q^{\top}$ and the $M \times M$ identity matrix, $\mathbf{I}_M$. Making use of \eqref{eq:kroneckerProduct}, this product becomes
\begin{equation}
	\mathbf{B}_q^{\top} \otimes \mathbf{I}_M = \left[
		\begin{array}{cccc}
			B_q(1,1) \mathbf{I}_M & B_q(2,1) \mathbf{I}_M & \cdots & B_q(N,1) \mathbf{I}_M \\
			B_q(1,2) \mathbf{I}_M & B_q(2,2) \mathbf{I}_M & \cdots & B_q(N,2) \mathbf{I}_M \\
			\vdots & \vdots & \ddots & \vdots \\
			B_q(1,L) \mathbf{I}_M & B_q(2,L) \mathbf{I}_M & \cdots & B_q(N,L) \mathbf{I}_M
		\end{array}
	\right],
\end{equation}
where $B_q(i,j)$ denotes the $(i,j)$-th element of matrix $\mathbf{B}_q$, given by \eqref{eq:matrixB}.
Hence, $\mathbf{B}_q^{\top} \otimes \mathbf{I}_M$ provides us with a matrix composed of $LN$ identity sub-matrices scaled by the corresponding coefficient of the sparse representation.
When this matrix is multiplied by $\boldsymbol\beta_q$ we obtain all the products between coefficients associated to nearby activations, thus increasing the value of the cost function when two such nearby activations occur.

\subsection{Sparse Solution through Successive Convex Approximations}
\label{sec:solutionIcassp2013}

Unfortunately, the new penalty term introduced in \eqref{eq:costFunctionModApprox} leads to a non-convex optimization problem.
However, in this subsection we present an algorithm, based on successive convex approximations (SCA) \citep{Chiang_Palomar_PowCont_GP_TWC2007,Wright_ApproxNonConvex_OpResearch1978,Avriel_AdvancesGP_1980}, for solving the constrained version of the Cross-Products LASSO.
In particular, the problem to be solved can be formulated alternatively as
\begin{equation} \label{eq:optimization}
  \begin{aligned}
& \underset{\boldsymbol\beta,\mathbf{c}}{\text{minimize}}
& &  \| \boldsymbol\beta \|_1 + \mathbf{c}^{\top} \boldsymbol\Gamma \mathbf{c} \\
& \text{subject to}
& & |\beta_k| = c_k, \qquad k=1,\ldots,MN \\
& & & \| \mathbf{y} - \boldsymbol\Phi \boldsymbol\beta  \| \leq \xi
\end{aligned}
\end{equation}
where $\beta_k$ (resp. $c_k$) is the $k$-th entry of $\boldsymbol\beta$ (resp $\mathbf{c}$), $\xi$ is some user-defined tolerable residual error, and the symmetric matrix $\boldsymbol\Gamma$, with zeros along its main diagonal, penalizes the cross products of the absolute values of $\boldsymbol\beta$. That is, the entry $\gamma_{k,\ell} \geq 0$ in the $k$-th row and $\ell$-th column of $\boldsymbol\Gamma$, induces a penalization $\gamma_{k,\ell} a_k a_{\ell} = \gamma_{k,\ell} |\beta_k| | \beta_{\ell} |$.

The optimization problem in \eqref{eq:optimization} is difficult to solve, since the cost function is not convex whenever $\boldsymbol\Gamma \ne \mathbf{0}$. Moreover, the first set of constraints is not convex. However, taking into account that the cost function increases with $c_k$, we notice that this problem is equivalent to
\begin{equation} \label{eq:optimization_2}
  \begin{aligned}
& \underset{\boldsymbol\beta,\bf{c}}{\text{minimize}}
& &  \mathbf{1}^{\top} \mathbf{c} + \mathbf{c}^{\top} \boldsymbol\Gamma \mathbf{c} \\
& \text{subject to}
& & |\beta_k| \leq c_k, \qquad k=1,\ldots,MN \\
& & & \| \mathbf{y} - \boldsymbol\Phi \boldsymbol\beta  \| \leq \xi
\end{aligned}
\end{equation}

Let us now introduce the constraint $\mathbf{1} + 2 \boldsymbol\Gamma \mathbf{c} \geq \mathbf{0}$, which ensures that, at the solution, the cost function increases with $c_k$. Although this constraint is redundant at this point, it will become relevant soon. Thus, the optimization problem is
\begin{equation} \label{eq:optimization_3}
  \begin{aligned}
& \underset{\boldsymbol\beta,\bf{c}}{\text{minimize}}
& &  \mathbf{1}^{\top} \mathbf{c} + \mathbf{c}^{\top} \boldsymbol\Gamma \mathbf{c} \\
& \text{subject to}
& & |\beta_k| \leq c_k, \qquad k=1,\ldots,MN \\
& & & \| \mathbf{y} - \boldsymbol\Phi \boldsymbol\beta  \| \leq \xi \\
& & & \mathbf{1} + 2 \boldsymbol\Gamma \mathbf{c} \geq \mathbf{0}
\end{aligned}
\end{equation}
where the main difficulty resides in the non-convex cost function. In order to deal with this difficulty and find a solution of the original Karush-Kuhn-Tucker (KKT) conditions \citep{Boyd_ConvexOpt}, we apply the SCA methodology \citep{Chiang_Palomar_PowCont_GP_TWC2007,Wright_ApproxNonConvex_OpResearch1978,Avriel_AdvancesGP_1980}. The main idea is replacing the non-convex functions by a sequence of local convex approximations, which must satisfy three conditions:
\begin{enumerate}
  \item The value of the original function, $f(\cdot)$, and its convex approximation, $\tilde{f}(\cdot)$, at the reference point
  	$\mathbf{x}_0$ should be the same, i.e., $f(\mathbf{x}_0) = \tilde{f}(\mathbf{x}_0)$.
  
  \item The gradients at the reference point should coincide, i.e., $\nabla_f(\mathbf{x}_0) = \nabla_{\tilde{f}}(\mathbf{x}_0)$.
  
  \item The convex approximation must be an over-estimator of $f(\cdot)$, i.e., $\tilde{f}(\mathbf{x}) \geq f(\mathbf{x})$, $\forall \mathbf{x}$.
\end{enumerate}
In our particular case, given a reference value $\mathbf{c}_0$ for the vector $\mathbf{c}$, the cost function can be approximated by $\mathbf{1}^{\top} \mathbf{c} + \mathbf{c}^{\top} \boldsymbol\Gamma_{+} \mathbf{c} + 2 \mathbf{c}_0^{\top} \boldsymbol\Gamma_{-} (\mathbf{c} - \mathbf{c}_0)$, where $\boldsymbol\Gamma_{+}$ and $\boldsymbol\Gamma_{-}$ are the positive semidefinite and negative semidefinite parts of $\boldsymbol\Gamma = \boldsymbol\Gamma_{+} + \boldsymbol\Gamma_{-}$. It is easy to check that this approximation satisfies the previous conditions, and therefore, the convex problem to be solved in each iteration of the proposed algorithm is finally,\footnote{Note that the constraint $\mathbf{1} + 2 \boldsymbol\Gamma \mathbf{c} \geq \mathbf{0}$ plays a crucial role in \eqref{eq:SCA}, ensuring that the first set of constraints is satisfied with equality $|\beta_k| = c_k$.}
\begin{equation} \label{eq:SCA}
  \begin{aligned}
& \underset{\boldsymbol\beta,\bf{c}}{\text{minimize}}
& &  \mathbf{1}^{\top} \mathbf{c} + \mathbf{c}^{\top} \boldsymbol\Gamma_{+} \mathbf{c} + 2 \mathbf{c}_0^{\top} \boldsymbol\Gamma_{-} (\mathbf{c} - \mathbf{c}_0) \\
& \text{subject to}
& & |\beta_k| \leq c_k, \qquad k=1,\ldots,MN \\
& & & \| \mathbf{y} - \boldsymbol\Phi \boldsymbol\beta  \| \leq \xi \\
& & & \mathbf{1} + 2 \boldsymbol\Gamma \mathbf{c} \geq \mathbf{0}
\end{aligned}
\end{equation}
The overall procedure is summarized in Algorithm \ref{Algorithm}, where the initial value for $\mathbf{c}$ ($\mathbf{c}_0 = \mathbf{0}$), reduces the cost function to the convex envelope of the original non-convex cost function.

\begin{algorithm}[!tb]
\begin{algorithmic}
\STATE{\textbf{Input}:$\boldsymbol\Gamma$, $\boldsymbol\Phi$, $\xi$ and $\mathbf{y}$.}
\STATE{\textbf{Output}: Recovered signal $\boldsymbol\beta$.}
\STATE{\textbf{Initialize} $\mathbf{c}_0 = \mathbf{0}$.}
\STATE{Obtain the Matrices $\boldsymbol\Gamma_{+}$ and $\boldsymbol\Gamma_{-}$ from the EV of $\boldsymbol\Gamma$}
\REPEAT
    \STATE{Solve the convex optimization problem in \eqref{eq:SCA}}
    \STATE{Update $\mathbf{c}_0 = \mathbf{c}$}
\UNTIL{Convergence}
\end{algorithmic}
\caption{SCA for Cross-Products LASSO} \label{Algorithm}
\end{algorithm}

\section{Sparse Spectral Analysis}
\label{sec:SSA}

\subsection{Iterative Deflation Approach for Spectral Analysis}
\label{sec:iterativeDeflationSSA}

Here we show the spectral analysis proposed on \citep{monzon_mlsp_2012}, which is based on applying an iterative deflation approach to the FFT of $\pi_q[n]$, extracting peaks with decreasing amplitudes up to a user defined threshold.
Hence, since we apply the spectral analysis to the inferred sparse activation sequence, we call our approach \emph{sparse spectral analysis} (SSA).
The number of peaks extracted (after the post-processing described in the following section) is an estimate of the number of existing foci and their locations provide us an estimate of their frequencies.\footnote{The location of the highest peak (i.e. the first one extracted) provides us with the dominant frequency.}

The first step in the SSA algorithm is segmenting $\pi_q[n]$ into $J$ windows containing $N_s = 4f_s$ samples (i.e. $\Lambda = 4$ s) without overlap, as done in DFA (see Section \ref{sec:DFA}).
Then we apply Algorithm \ref{alg:spectralAnalysis} to the FFT of each segment, $\Pi_q^j(f) = \mathcal{F} \left\lbrace \pi_q^j[n] \right\rbrace $ with $1 \le j \le J$, after bandpass filtering.
Algorithm \ref{alg:spectralAnalysis} follows a \emph{deflation approach}, searching iteratively for the highest peak of $|\Pi_q^j(f)|$ within the frequency range that is physiologically interpretable ($0.5 \le f_r \le 2$ Hz for sinus rythm and $2 \le f_r \le 10$ Hz for AF) and adding it to the set of potential activation frequencies, $\hat{\mathbf{f}}_q^{j}$.
After each iteration we apply a second-order IIR digital notch filter to the signal centered around the detected frequency with bandwith $B_{\text{3 dB}} = 2f_{\Lambda} = 0.5$ Hz to eliminate the detected peak before searching for a new one.
The algorithm stops when the highest peak detected is below a threshold, $\Gamma_q^j = \gamma_q \max |\Pi_q^j(f)|$, being $\gamma_q$ a user defined parameter.

\begin{algorithm}[!htb]
	\begin{algorithmic}	
		\FOR {j=1 to J}
			\STATE Initialize $\Gamma_q^j = \gamma_q \cdot \max \vert \Pi_q^j(f)\vert $.
			\STATE Initialize $i=1$ and $\pi_{q,1}^j(t) = \pi_q^j (t)$
			\WHILE {$\max \vert \Pi_{q,i}^j(f) \vert \ge \Gamma_q^j$}				
				\STATE 1. Calculate the spectrum: $ \Pi_{q,i}^j(f) = \mathcal{F} \left\lbrace \pi_{q,i}^{j}(t)\right\rbrace $ 				
				\STATE 2. Obtain $\hat{f}_q^{j}(i) = \argmax{f} \vert \Pi_{q,i}^j(f) \vert$
				\STATE 3. Filter the signal: $\pi_{q,i+1}^j(t) = \pi_{q,i}^j(t) * h_{notch}(t) $				
				\STATE 4. $i = i+1$		
			\ENDWHILE
		\ENDFOR
	\end{algorithmic}	
	\caption{Iterative Spectral Analysis for the $q$-th signal.}
	\label{alg:spectralAnalysis}
\end{algorithm}

\begin{figure}[!htb]
\begin{minipage}[b]{1.0\linewidth}
  \centering
  \centerline{\includegraphics[width=0.7\textwidth]{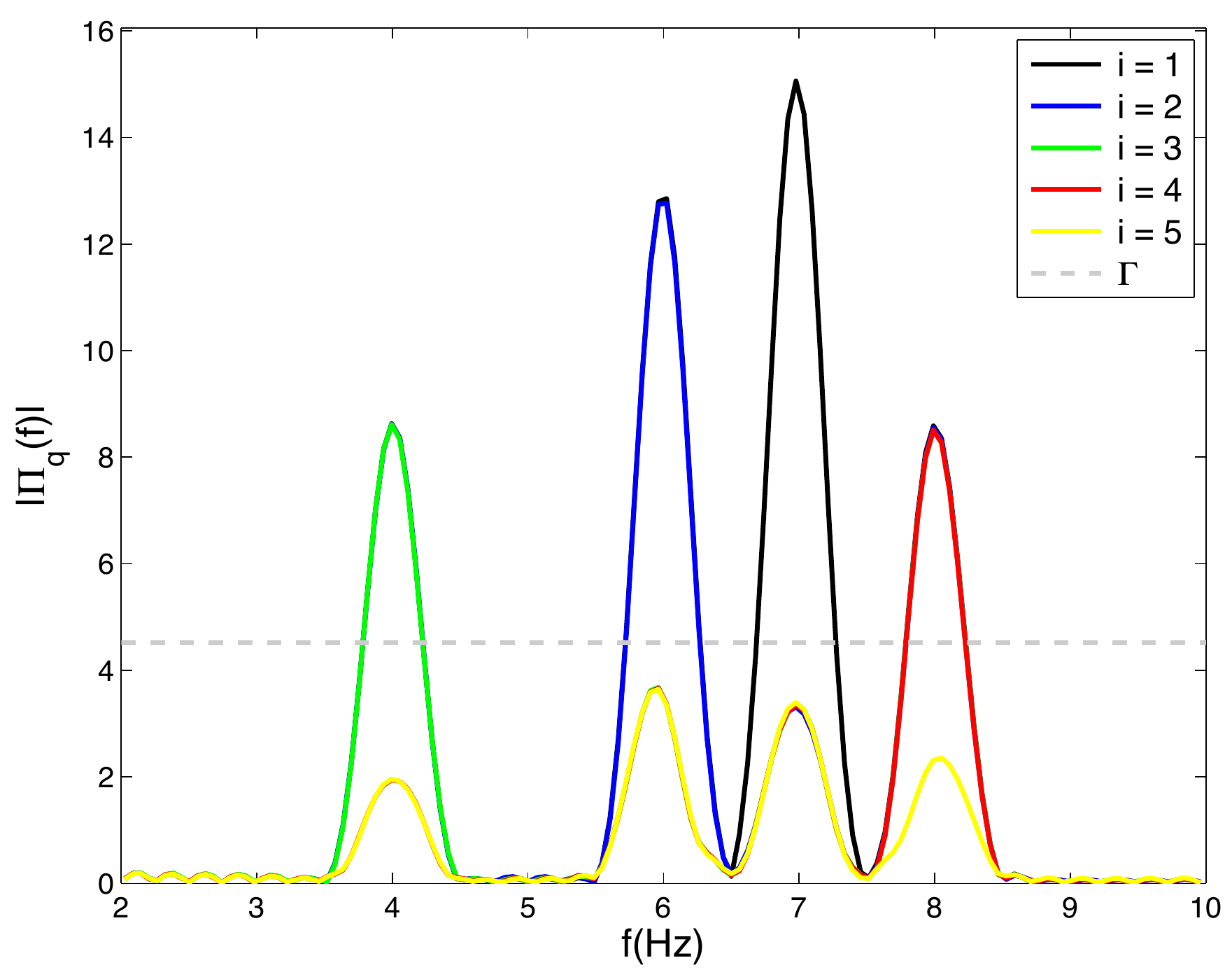}} 
\end{minipage}
\caption{\small{Example of the SSA for a single segment of $\pi_q[n]$}.}
\label{fig:algExample}
\end{figure}

Figure \ref{fig:algExample} shows an example of the spectrum obtained iteratively for a single segment.
The activation sequence, $\pi_q[n]$, has been syntheticaly generated using $R=3$ foci (with $f_1 = 4$ Hz, $f_2 = 6$ Hz and $f_3 = 7$ Hz) and random phases.
The highest peak for the amplitude spectrum in the first iteration (shown in black) is $\hat{f}_1 \approx 6.98$ Hz, which corresponds to $f_3$.
Then we apply the notch filter centered around $\hat{f}_1$ to the signal, obtaining the amplitude spectrum shown in blue, and detecting $\hat{f}_2 \approx 5.96$ Hz, which is close to $f_2$.
After a second notch filtering centered around $\hat{f}_2$, the third iteration (in green) detects $\hat{f}_3 \approx 4$ Hz, which corresponds to $f_1$.
After notch filtering again, iteration 4 (in red) detects $\hat{f}_4 \approx 7.99$ Hz, which is the first harmonic of $\hat{f}_3$.
Finally, after another notch filtering, all the peaks of the spectrum in the fifth iteration (in yellow) fall below the threshold $\Gamma_q = 0.3 \times \max \vert \Pi_q(f) \vert$. Hence, the algorithm concludes after obtaining 4 potential frequencies: $\hat{\mathbf{f}}_q = [6.98, 5.96, 4, 7.99]$ Hz.
%

\subsection{Post-Processing: Discarding Harmonics}
\label{sec:postProcessing}

The post-processing stage takes the set of potential activation frequencies detected inside each window, $\hat{\mathbf{f}}_q^j$, and determines whether they belong to different activation foci or not applying the following steps:
\begin{enumerate}
	\item Elimination of repeated frequencies. Two frequencies, $f_1$ and $f_2$, correspond to the same focus if $ \vert f_1 - f_2 \vert 			\leq f_{\Lambda}$. If this happens, the one associated to the smallest peak is deleted.
	\item Analysis of 2/3 frequency relationships. Due to the frequency range used in the analysis, given a single frequency, $f_0$, in
		practice we can find at most two harmonics: $f_1 = 2f_0$ and $f_2 = 3f_0$. Thus, if we have detected the first and second harmonic 
		of a given frequency, $f_0$, their relationship will be $f_1 = \frac{2}{3} f_2$. Here we check this relationship, keeping only the
		frequency associated to a higher amplitude in the spectrum when we find it.
	\item Discovery of harmonics and subharmonics. When two detected frequencies have a harmonic or subharmonic relationship, we only
		keep the one detected first in the spectral analysis and deleting the other.
	\item Discovery of cross-modulation frequencies. We analyze whether each new element in $\hat{\mathbf{f}}_q$ is a cross-modulation
		product of two previously detected frequencies, i.e. whether $f_3 = \pm mf_1 \pm nf_2$ for any two integers $m$ and $n$. In
		this case $f_3$ will be deleted.
\end{enumerate}

With this analysis, we are able to estimate the number of activation foci present in our EGMs, $\hat{R}_q^j$, as well as their frequencies, $\hat{\mathbf{f}}_q^j$.
%
Continuing with the example shown in Figure \ref{fig:algExample}, the post-processing will find out that $\hat{f}_4$ is the first harmonic of $\hat{f}_3$, deleting it and obtaining a correct final estimation of $\hat{R} = 3$ activation foci with frequencies $\hat{\mathbf{f}}_q = [6.98, 5.95, 4]$ Hz, which are quite close to the true ones.
%
%

\subsection{Alternative Spectral Analysis based on Eigen-Values}
\label{sec:eigenValueSSA}

We are currently considering alternative SSA approaches using methods based on eigen-values, such as ROOT MUSIC.
%



\section{Conclusions and Future Lines}
\label{sec:conclusions}

Contributions of the technical report:
\begin{enumerate}
	\item New more realistic mathematical model for EGM signals introduced based on latent signals (sparse activations or spike trains).
	\item Sparse reconstruction model based on an overcomplete dictionary proposed.
	\item Examples of dictionary construction based on Hermitian wavelets of order zero (energy-normalized Gaussians) and order one
		(Mexican hat wavelets).
	\item Indirect sparse solution of the problem using a LASSO regularization initially followed by a second stage to enforce the
		biological restrictions imposed by the refractory period of cardiac cells.
	\item Direct sparse solution of the problem using a new (non-convex) regularization term that we call cross-products LASSO (CP-LASSO).
	\item Successive convex approximations (SCA) approach introduced for solving the non-convex CP-LASSO optimization problem.
\end{enumerate}
Future lines:
\begin{enumerate}
	\item Perform many more simulations on real EGM data.
	\item Extend to the multi-channel case, probably using some type of Group LASSO formulation, although the way in which the groups
		are defined/learnt is unclear yet.
\end{enumerate}

\appendix

\section*{Derivation of the Discrete-Time Convolutional Model}
\label{sec:convolution}

Since the sparse approximation is applied on the discrete-time difference sequence, $z_q[n]$, the coefficients $\beta_{m,q}[n]$ will have the same support as this sequence, i.e., $1 \le n \le N$.
Hence, $\beta_{m,q}[n]$ may be expressed as
\begin{equation}
	\beta_{m,q}[n] = \sum_{k=1}^{N}{\beta_{m,q}[k] \delta[n-k]}
		= \beta_{m,q}[n] (u[n-1] - u[n-(N+1)]),
\label{eq:coefficientSequence}
\end{equation}
where $\delta[n]$ denotes Kronecker's delta and $u[n]$ Heaviside's unit step function.
Regarding the elements of the dictionary, since the unknown input-output channels, $h_{r,q}[n]$, are assumed to be causal, here we will always consider causal discrete-time activations, $G_m[n]$, typically obtained from a non-causal waveform, $G_m(t)$ with support $-T_M \le t \le T_M$, through sampling and time-shifting.
Thus, the support for $G_m[n]$ will be $0 \le n \le 2N_M$ (with $N_M = \lfloor T_M/T_s \rfloor$, where $\lfloor x \rfloor$ denotes the integer part of $x$, indicating the last non-zero element in the discrete-time waveform before time-shifting), and this sequence may be expressed as
\begin{equation}
	G_m[n] = \sum_{k=0}^{2N_M}{G_m[k] \delta[n-k]}
		= G_m[n] (u[n] - u[n-(2N_M+1)]).
\label{eq:activationSequence}
\end{equation}

Now we can formulate the convolution in \eqref{eq:diff_model} as
\begin{equation}
	\beta_{m,q}[n] * G_m[n] = \sum_{k=-\infty}^{\infty}{\beta_{m,q}[k](u[k-1]-u[k-(N+1)])G_m[n-k](u[n-k]-u[n-k-(2N_M+1)])}.
\label{eq:convolution1}
\end{equation}
In order to establish the limits for this convolution, we notice that
\begin{equation}
	u[k-1]-u[k-(N+1)] \ne 0 \Leftrightarrow
		\begin{cases}
			k-1 \ge 0 \Rightarrow k \ge 1,\\
			k-(N+1) < 0 \Rightarrow k < N+1 \Rightarrow k \le N,
		\end{cases}
\label{eq:limits1}
\end{equation}
\begin{equation}
	u[n-k]-u[n-k-(2N_M+1)] \ne 0 \Leftrightarrow
		\begin{cases}
			n-k \ge 0 \Rightarrow k \le n,\\
			n-k-(2N_M+1) < 0 \Rightarrow k > n-(2N_M+1) \Rightarrow k \ge n-2N_M.
		\end{cases}
\label{eq:limits2}
\end{equation}
Therefore, the lower limit for the convolution will be
\begin{equation}
	k_{inf} = \sup\{1,n-2N_M\} = 1 \vee (n-2N_M),
\label{eq:lowerLimit1}
\end{equation}
whereas the upper limit will be
\begin{equation}
	k_{sup} = \inf\{n,N\} = n \wedge N = n.
\label{eq:upperLimit1}
\end{equation}
Inserting these limits in \eqref{eq:convolution1}, we obtain
\begin{equation}
	\beta_{m,q}[n] * G_m[n] = \sum_{k=1 \vee (n-2N_M)}^{n}{\beta_{m,q}[k]G_m[n-k]}.
\label{eq:convolution2}
\end{equation}
Finally, by stating explicitly that $G_m[n] = 0$ for $n > 2N_M$ and $n < 0$, we may remove the supremum from the lower limit and perform the sum from 1 up to $N$, as is done in \eqref{eq:diff_model}, although there will only be at most $2N_M+1$ non-zero terms in the sum.

Alternatively, we can formulate the convolution in \eqref{eq:diff_model} as
\begin{equation}
	\beta_{m,q}[n] * G_m[n] = \sum_{k=-\infty}^{\infty}{G_m[k](u[k]-u[k-(2N_M+1)])\beta_{m,q}[n-k](u[n-k-1]-u[n-k-(N+1)])}.
\label{eq:convolution3}
\end{equation}
In order to establish the limits for this convolution, we notice again that
\begin{equation}
	u[k]-u[k-(2N_M+1)] \ne 0 \Leftrightarrow
		\begin{cases}
			k \ge 0,\\
			k-(2N_M+1) < 0 \Rightarrow k < 2N_M+1 \Rightarrow k \le 2N_M,
		\end{cases}
\label{eq:limits3}
\end{equation}
\begin{equation}
	u[n-k-1]-u[n-k-(N+1)] \ne 0 \Leftrightarrow
		\begin{cases}
			n-k-1 \ge 0 \Rightarrow k \le n-1,\\
			n-k-(N+1) < 0 \Rightarrow k > n-(N+1) \Rightarrow k \ge n-N.
		\end{cases}
\label{eq:limits4}
\end{equation}
Therefore, now the lower limit for the convolution will be
\begin{equation}
	k_{inf} = \sup\{0,n-N\} = 0 \vee (n-N) = 0,
\label{eq:lowerLimit2}
\end{equation}
whereas the upper limit will be
\begin{equation}
	k_{sup} = \inf\{2N_M,n-1\} = (2N_M) \wedge (n-1).
\label{eq:upperLimit2}
\end{equation}
Inserting these limits in \eqref{eq:convolution1}, we obtain
\begin{equation}
	\beta_{m,q}[n] * G_m[n] = \sum_{k=0}^{(2N_M) \wedge (n-1)}{G_m[k]\beta_{m,q}[n-k]}.
\label{eq:convolution4}
\end{equation}
Finally, by stating explicitly that $G_m[n] = 0$ for $n > 2N_M$ and $\beta_{m,q}[n] = 0$ for $n < 1$, we may remove the infimum from the upper limit and perform the sum from 0 up to $N-1$, although there will only be at most $2N_M+1$ non-zero terms in the sum again.

\section*{Derivation of the Hermitian Dictionaries}
\label{sec:hermitian}

In this section we derive the hermitian wavelet dictionaries used for the sparse reconstruction.
First we obtain the normalization factor for the Gaussian function, which corresponds to the zero-th order Hermitian wavelet.
Then we develop the normalized expressions for the first and second order Hermitian wavelets.
Finally, we briefly discuss the general shape of the $\ell$-th order Hermitian wavelet.

\subsection*{Zero-th Order Hermitian Wavelet (Energy-Normalized Gaussian)}
\label{sec:normalizedGaussian}

Let us denote the $m$-th standard Gaussian function as
\begin{equation}
	\widetilde{\phi}_m^{(0)}(t) = \frac{1}{\sqrt{2\pi\sigma_m^2}} \exp\left(-\frac{t^2}{2\sigma_m^2}\right).
\label{eq:standardGaussian}
\end{equation}
This function corresponds to a proper and normalized probability density function (PDF), i.e. $\widetilde{\phi}_m^{(0)}(t) \ge 0$ for $-\infty < t < \infty$, and
\begin{equation}
	\int_{-\infty}^{\infty}{\widetilde{\phi}_m^{(0)}(t) dt} = 1.
\end{equation}
However, it is not normalized in energy, since
\begin{align}
	E_m^{(0)} = \| \widetilde{\phi}_m^{(0)}(t) \|_2^2 & = \int_{-\infty}^{\infty}{|\widetilde{\phi}_m^{(0)}(t)|^2 dt} \nonumber \\
		& = \frac{1}{2\pi\sigma_m^2} \int_{-\infty}^{\infty}{\exp\left(-\frac{t^2}{\sigma_m^2}\right) dt} \nonumber \\
		& = \frac{\sqrt{2\pi(\sigma_m/\sqrt{2})^2}}{2\pi\sigma_m^2} \mathbb{E}\{t^0\} = \frac{1}{2\sqrt{\pi}\sigma_m}
\label{eq:energy0}
\end{align}
is not equal to one unless we have $\sigma_m = 1/[2\sqrt{\pi}]$.
In the sequel we will use $\mathbb{E}\{f(t)\}$ to denote the expectation of $f(t)$ w.r.t. a Gaussian centered around the origin with standard deviation $\sigma_m/\sqrt{2}$, i.e.
\begin{align}
	\mathbb{E}\{f(t)\} & = \int_{-\infty}^{\infty}{\frac{f(t)}{\sqrt{2\pi(\sigma_m/\sqrt{2})^2}}
			\exp\left(-\frac{t^2}{2(\sigma_m/\sqrt{2})^2}\right) dt} \nonumber \\
		& = \frac{1}{{\sqrt{\pi\sigma_m^2}}} \int_{-\infty}^{\infty}{f(t) \exp\left(-\frac{t^2}{\sigma_m^2}\right) dt}.
\label{eq:expectation}
\end{align}
Finally, making use of \eqref{eq:standardGaussian} and \eqref{eq:energy0}, the energy-normalized version of the Gaussian function will be
\begin{equation}
	\phi_m^{(0)}(t) = \frac{\widetilde{\phi}_m^{(0)}(t)}{\sqrt{\|\widetilde{\phi}_m^{(0)}(t)\|_2^2}}
		= \frac{1}{\pi^{1/4}\sqrt{\sigma_m}} \exp\left(-\frac{t^2}{2\sigma_m^2}\right),
\label{eq:normalizedGaussian}
\end{equation}
which is precisely the expression given by \eqref{eq:gauss_base}.

\subsection*{First-Order Hermitian Wavelet}
\label{sec:Hermitian1}

The first-order hermitian wavelet is the negative normalized first derivative of the Gaussian function.
Taking the first derivative of \eqref{eq:standardGaussian} we obtain the following unnormalized function
\begin{equation}
	\widetilde{\phi}_m^{(1)}(t) = \frac{d\widetilde{\phi}_m^{(0)}(t)}{dt}
		= -\frac{t}{\sqrt{2\pi}\sigma_m^3} \exp\left(-\frac{t^2}{2\sigma_m^2}\right),
\label{eq:unnormalizedHermitian1}
\end{equation}
which has an energy
\begin{align}
	E_m^{(1)} = \| \widetilde{\phi}_m^{(1)}(t) \|_2^2 & = \int_{-\infty}^{\infty}{|\widetilde{\phi}_m^{(1)}(t)|^2 dt} \nonumber \\
		& = \frac{1}{2\pi\sigma_m^6} \int_{-\infty}^{\infty}{t^2 \exp\left(-\frac{t^2}{\sigma_m^2}\right) dt} \nonumber \\
		& = \frac{\sqrt{\pi\sigma_m^2}}{2\pi\sigma_m^6} \mathbb{E}\{t^2\} = \frac{1}{4\sqrt{\pi}\sigma_m^3},
\label{eq:energy1}
\end{align}
where we have used the fact that $\mathbb{E}\{t^2\} = \sigma_m^2/2$, following the definition of the expectation operator provided by \eqref{eq:expectation}.
Finally, making use of \eqref{eq:unnormalizedHermitian1} and \eqref{eq:energy1}, the energy-normalized first-order hermitian wavelet will be
\begin{equation}
	\phi_m^{(1)}(t) = -\frac{\widetilde{\phi}_m^{(1)}(t)}{\sqrt{\|\widetilde{\phi}_m^{(1)}(t)\|_2^2}}
		= \frac{2t}{\pi^{1/4}\sqrt{2\sigma_m^3}} \exp\left(-\frac{t^2}{2\sigma_m^2}\right).
\label{eq:normalizedHermitian1}
\end{equation}

\subsection*{Second-Order Hermitian Wavelet (Mexican Hat Wavelet)}
\label{sec:Hermitian2}

The second-order hermitian wavelet is the negative normalized second derivative of the Gaussian function.
Taking the first derivative of \eqref{eq:unnormalizedHermitian1} we obtain the following unnormalized function
\begin{align}
	\widetilde{\phi}_m^{(2)}(t) = \frac{d\widetilde{\phi}_m^{(1)}(t)}{dt}
		& = -\frac{1}{\sigma_m^2} \left(\widetilde{\phi}_m^{(0)}(t) + t \widetilde{\phi}_m^{(1)}(t)\right) \nonumber \\
		& = -\frac{1}{\sqrt{2\pi}\sigma_m^3} \left(1-\frac{t^2}{\sigma_m^2}\right) \exp\left(-\frac{t^2}{2\sigma_m^2}\right),
\label{eq:unnormalizedHermitian2}
\end{align}
which has an energy
\begin{align}
	E_m^{(2)} = \| \widetilde{\phi}_m^{(2)}(t) \|_2^2 & = \int_{-\infty}^{\infty}{|\widetilde{\phi}_m^{(2)}(t)|^2 dt} \nonumber \\
		& = \frac{1}{2\pi\sigma_m^6}
			\int_{-\infty}^{\infty}{\left(1-\frac{t^2}{\sigma_m^2}\right)^2 \exp\left(-\frac{t^2}{\sigma_m^2}\right) dt} \nonumber \\
		& = \frac{\sqrt{\pi\sigma_m^2}}{2\pi\sigma_m^6}
			\left[\mathbb{E}\{t^0\} - \frac{2}{\sigma_m^2} \mathbb{E}\{t^2\} + \frac{1}{\sigma_m^4} \mathbb{E}\{t^4\}\right] \nonumber \\
		& = \frac{1}{2\sqrt{\pi}\sigma_m^5}
			\left[1 - \frac{2}{\sigma_m^2}\frac{\sigma_m^2}{2} + \frac{1}{\sigma_m^4}\frac{3\sigma_m^4}{4}\right]
			= \frac{3}{8\sqrt{\pi}\sigma_m^5},
\label{eq:energy2}
\end{align}
where we have used the fact that $\mathbb{E}\{t^4\} = 3\sigma_m^4/4$, following the definition of the expectation operator provided by \eqref{eq:expectation}.
Finally, making use of \eqref{eq:unnormalizedHermitian2} and \eqref{eq:energy2}, the energy-normalized second-order hermitian wavelet, also known as Mexican hat or Ricker wavelet, will be
\begin{equation}
	\phi_m^{(2)}(t) = -\frac{\widetilde{\phi}_m^{(2)}(t)}{\sqrt{\|\widetilde{\phi}_m^{(2)}(t)\|_2^2}}
		= \frac{2}{\pi^{1/4}\sqrt{3\sigma_m}} \left(1-\frac{t^2}{\sigma_m^2}\right) \exp\left(-\frac{t^2}{2\sigma_m^2}\right).
\label{eq:normalizedHermitian2}
\end{equation}

\subsection*{Higher-Order Hermitian Wavelets}
\label{sec:HermitianGeneral}

In general, the $\ell$-th order Hermitian wavelet, $\phi_m^{(\ell)}(t)$ for $\ell \ge 1$, is obtained as the $\ell$-th negative normalized derivative of the Gaussian function.
In order to derive $\phi_m^{(\ell)}(t)$ we follow the three-step procedure used for the first and second order wavelets: we calculate first the unnormalized derivative,
\begin{equation}
	\widetilde{\phi}_m^{(\ell)}(t) = \frac{d^{\ell}\widetilde{\phi}_m^{(0)}(t)}{dt^{\ell}}
		= \frac{d^{\ell-1}\widetilde{\phi}_m^{(1)}(t)}{dt^{\ell-1}} = \cdots
		= \frac{d\widetilde{\phi}_m^{(\ell-1)}(t)}{dt},
\label{eq:unnormalizedHermitianGeneral}
\end{equation}
which can be obtained easily by applying the following recursion,\footnote{The recursion in \eqref{eq:recursionHermitian} is valid for $\ell \ge 2$ and can be easily proved by induction. Indeed, by defining $\widetilde{\phi}_m^{(-1)}(t) = 0$, it is valid even for $\ell=1$.}
\begin{equation}
	\widetilde{\phi}_m^{(\ell)}(t) = -\frac{1}{\sigma_m^2}
		\left((\ell-1) \widetilde{\phi}_m^{(\ell-2)}(t) + t \widetilde{\phi}_m^{(\ell-1)}(t)\right);
\label{eq:recursionHermitian}
\end{equation}
then we obtain its energy,
\begin{equation}
	E_m^{(\ell)} = \| \widetilde{\phi}_m^{(\ell)}(t) \|_2^2 = \int_{-\infty}^{\infty}{|\widetilde{\phi}_m^{(\ell)}(t)|^2 dt},
\label{eq:energyGeneral}
\end{equation}
and the $\ell$-th order normalized Hermitian wavelet is finally given by
\begin{equation}
	\phi_m^{(\ell)}(t) = -\frac{\widetilde{\phi}_m^{(\ell)}(t)}{\sqrt{\|\widetilde{\phi}_m^{(\ell)}(t)\|_2^2}}.
\label{eq:normalizedHermitianGeneral}
\end{equation}
Following this procedure, we notice that the $\ell$-th order Hermitian wavelet can be expressed as
\begin{equation}
	\widetilde{\phi}_m^{(\ell)}(t) = \kappa_m^{(\ell)} H_{\ell}\left(\frac{t}{\sqrt{2\sigma_m^2}}\right)
		\exp\left(-\frac{t^2}{2\sigma_m^2}\right),
\label{eq:normalizedHermitianGeneral2}
\end{equation}
where $H_{\ell}(x)$ denotes the $\ell$-th order Hermite polynomial \citep{Abramowitz:handbookMathFcn1965} and $\kappa_m^{(\ell)}$ is a normalization constant, obtained as
\begin{align}
	\kappa_m^{(\ell)} & = \left[ \int_{-\infty}^{\infty}{\left(H_{\ell}\left(\frac{t}{\sqrt{2\sigma_m^2}}\right)\right)^2
			\exp\left(-\frac{t^2}{2\sigma_m^2}\right) dt}\right]^{-1/2} \nonumber \\
		& = \left[\sqrt{\pi\sigma_m^2}
			\mathbb{E}\left\{\left(H_{\ell}\left(\frac{t}{\sqrt{2\sigma_m^2}}\right)\right)^2\right\} \right]^{-1/2},
\label{eq:normalizationConstant}
\end{align}
where the last expectation is as defined in \eqref{eq:expectation}.


\bibliographystyle{IEEEbib}
\bibliography{AF,AFtechRep,via}

\end{document}